\documentclass[11pt]{article}
\usepackage[margin=1in]{geometry}
\usepackage[T1]{fontenc}
\usepackage{lmodern}
\usepackage{microtype}
\usepackage{hyperref}
\usepackage[title]{appendix}
\usepackage{mdframed}
\usepackage{graphicx}
\usepackage{amssymb}
\usepackage{amsmath}
\usepackage{xcolor}
\usepackage{pdfpages}

\hypersetup{
  pdftitle={An N-Plus-1 GPT Agency for Critical Solution of Mechanical Engineering Analysis Problems},
  pdfauthor={Anthony T. Patera and Rohan C. Abeyaratne},
  pdfkeywords={multi-agent systems, generative AI, engineering analysis, probabilistic reliability, mechanical engineering}
}
 
\title{An $N$-Plus-1 GPT Agency for Critical Solution of\\ Mechanical Engineering Analysis Problems}
 
\author{Anthony T. Patera \\
Department of Mechanical Engineering, Massachusetts Institute of Technology \\
\texttt{patera@mit.edu}
\and
Rohan C. Abeyaratne \\
Department of Mechanical Engineering, Massachusetts Institute of Technology \\
\texttt{rohan@mit.edu}
}
 
\date{\today}
 
\begin{document}
\maketitle

\begin{abstract}
Generative AI, and specifically GPT, can produce a remarkable solution to a mechanical engineering analysis problem --- but also, on occasion, a flawed solution. For example, an elementary mechanics problem is solved flawlessly in one GPT instance and incorrectly in a subsequent GPT instance, with a success probability of only $\sim$85\%. This unreliability renders ``out-of-the-box'' GPT unsuitable for deployment in education or engineering practice. 
We introduce an ``$N$-Plus-1'' GPT Agency for Initial (Low-Cost) Analysis of mechanical engineering Problem Statements. Agency first  launches $N$ instantiations of Agent Solve to yield $N$ independent Proposed Problem Solution Realizations; Agency then invokes Agent Compare to summarize and compare the $N$ Proposed Problem Solution Realizations and to provide a Recommended Problem Solution. We argue from Condorcet’s Jury Theorem that, for a Problem Statement characterized by per-Solve success probability greater than $1/2$ (and $N$ sufficiently large), the Predominant (Agent Compare) Proposed Problem Solution will, with high probability, correspond to a Correct Proposed Problem Solution. Furthermore, Agent Compare can also incorporate aspects of Secondary (Agent Compare) Proposed Problem Solutions, in particular when the latter represent alternative  Problem Statement interpretations  --- different Mathematical Models --- or alternative Mathematical Solution Procedures. Comparisons to Grok Heavy, a commercial multi-agent model, show similarities in design and performance, but also important differences in emphasis: our Agency focuses on transparency and pedagogical value.\end{abstract}

\section*{Acknowledgements}
We would like to thank Dr.\ Kento Kaneko for his many contributions to this work: beneficial discussions on pedagogy, general software consultations, and API code for image processing. We would also like to thank Professor David Parks for enlightening discussions. Finally, we are grateful to David Biglen for alerting us to Grok recent developments.

ATP would like to thank the MIT Department of Mechanical Engineering for Professor-Post-Tenure support and also OpenAI API resources.

\section{Motivation} \label{sec:motivation}

\begin{figure}[!ht]
\begin{mdframed}
\includegraphics[width = 0.988 \textwidth]{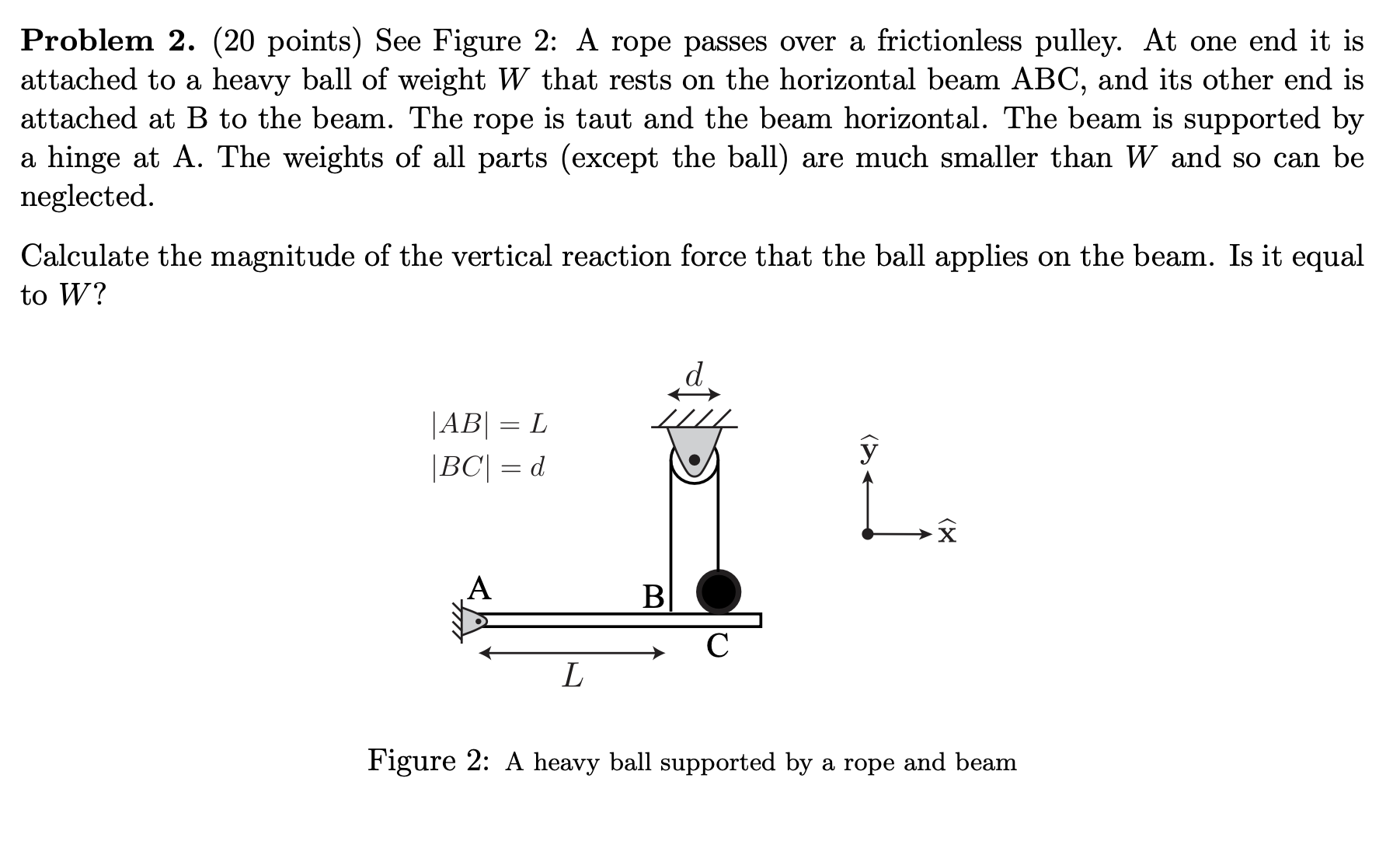}
\end{mdframed}
\caption{\label{fig:PROBBallBeam_def} Problem Statement (.png file) for PROBBallBeam.}
\end{figure}

We consider in Figure \ref{fig:PROBBallBeam_def} a problem from MIT Mechanical Engineering subject 2.001, {\em Mechanics and Materials I}: Problem Statement PROBBallBeam. PROBBallBeam is ostensibly simple but also subtle. 

We apply GPT Model \texttt{o4-mini}, \texttt{reasoning\_effort=high} to the PROBBallBeam .png file of Figure \ref{fig:PROBBallBeam_def}. We obtain a flawless Proposed Problem Solution, Correct and also well-argued: the wonder of Generative AI. We repeat the experiment --- a second, independent, realization. We obtain now a seriously flawed (Incorrect) Proposed Problem Solution: the frustration of Generative AI. A more systematic study reveals that the probability of success --- the probability that GPT Model \texttt{o4-mini}, \texttt{reasoning\_effort=high} returns a Correct Proposed Problem Solution to PROBBallBeam ---  is approximately 0.85. {\em Not nearly large enough to justify GPT deployment ``out-of-the-box'' either in education or in engineering practice.}

However, in a properly conceived GPT Agency  (a collaboration of several GPT Agents) we can transform GPT from a truly remarkable but erratic novelty into a truly capable and reliable assistant. In this work, we propose an ``$N$-Plus-$1$'' GPT Agency for critical solution of Mechanical Engineering analysis problems.

\section{Overview} \label{sec:overview}

\noindent \textbf{Motivation}: Generative AI, and specifically GPT, can produce a remarkable solution to a mechanical engineering analysis problem --- but also, on occasion, a flawed solution. For example, the simple-but-subtle ``Ball and Beam'' problem (MIT 2.001) of Figure \ref{fig:PROBBallBeam_def} is solved flawlessly in a first GPT instance and incorrectly in a second instance, with an overall success probability of only $\sim$85\%. This unreliability renders ``out-of-the-box'' GPT unsuitable for deployment in education or engineering practice. 

The central hypothesis of this paper is that reliability can be dramatically improved through an ``$N$-Plus-1'' GPT Agency---a structured system of multiple GPT agents. $\Box$

\medskip

\noindent \textbf{Framework: Problem Statements and Problem Solutions}:  
We define a structured framework:  

\begin{itemize}
\item {\em Problem Statements} include natural language and images specifying the artifact, operating conditions, and quantities of interest (QoI).  
\item {\em Problem Solutions} progress through four stages: (1) Data completion, (2) Development of a mathematical model, (3) Development of a mathematical solution procedure, and (4) Verification and validation.  
\end{itemize}

Our focus is on Initial Analysis---low-cost, approximate solutions that are prevalent in engineering education and often sufficient for conceptual design and early-stage decision-making. $\Box$

\medskip

\noindent \textbf{The $N$-Plus-1 GPT Agency}:  
The Agency is a collection of GPT agents orchestrated to solve problems collaboratively. The key agents are Agent Solve and Agent Compare:

\begin{itemize}
\item {\em Agent Solve}: Generates $N$  independent candidate solutions, denoted Proposed Problem Solution Realizations.  
\item {\em Agent Compare}: Reviews and synthesizes the candidate solutions to provide User with a critical comparison as well as a Recommended Problem Solution (with justification).  
\end{itemize}
In tests on problem sets from MIT Mechanical Engineering subjects 2.001 (mechanics) and 2.51 (heat transfer), Agency results significantly outperformed single GPT runs. $\Box$

\medskip

\noindent \textbf{Desired Behaviors and Benefits}:  Agency $N$-Plus-1 is designed to serve four types of Users: Practitioners (engineers), (Agency software) Developers, Students, and Instructors. Key desired outcomes include:  

\begin{itemize}
\item For Practitioners and Students: saving time, reducing drudgery, increasing reliability, and {\em ensuring effective critical engagement with AI-generated solutions}.  
\item For Developers: monitoring performance as new GPT models are released.  
\item For Instructors: enabling scalable student access to advanced GPT models, supporting (improved) problem design, and assisting in Student assessment. 
\end{itemize}

In education, the aim is to shift Student responsibility from generating Proposed Problem Solution to effectively (efficiently) verifying, critiquing, and improving AI-generated  Proposed Problem Solutions. We do not intend Agency $N$-Plus-1 as a tutor; rather, we intend Agency $N$-Plus-1 as a problem solver---Agency $N$-Plus-1 proposes a Recommended Problem Solution, and the Practitioner or Student accepts or rejects. $\Box$

\medskip

\noindent \textbf{Probabilistic Foundation}:  We define the (empirical) Prevalent Proposed Problem Solution (or Prevalent Opinion) as the mode --- most frequent --- Proposed Problem Solution in our sample of $N$ Proposed Problem Solution Realizations (ties resolved by a fixed rule); a Prevalent Opinion is denoted (empirical) Predominant if there are greater than $N/2$ occurrences of the Prevalent Opinion. We denote by Secondary Proposed Problem Solutions (or Secondary Opinions) the remaining Proposed Problem Solutions --- non-Prevalent Opinions.
 
We argue from Condorcet’s Jury Theorem  \cite{condorcet1785} (see also \cite{meir2022} for recent embodiments) that, for a Problem Statement characterized by per-Solve success probability greater than $1/2$ (and $N$ sufficiently large), the Predominant (Agent Compare) Proposed Problem Solution will, with high probability, correspond to a Correct Proposed Problem Solution. (Note that, as the success probability approaches 1, the requisite $N$ will decrease.) Furthermore, Agent Compare can feature not only the Predominant or Prevalent Opinion, but also incorporate aspects of Secondary (Agent Compare) Proposed Problem Solutions, in particular when the latter represent alternative  Problem Statement interpretations  --- different Mathematical Models --- or alternative Mathematical Solution Procedures; Secondary Opinions are particularly important in the case of epistemic uncertainty (multiple valid solutions) or missing physics (systematic blind spots). This approach transforms AI’s variability from a liability into an asset, using diversity of responses to reveal uncertainty, error, or alternative modeling assumptions. $\Box$

\medskip

\noindent \textbf{Novelty}:  Commercial embodiment of related multi-agent ideas is found in Grok Heavy \cite{musk2025}. Our approach shares some aspects with Grok Heavy---notably, multiple parallel agents and synthesis of a final recommendation---but differs in important respects. Grok agents collaborate in a largely opaque fashion, with convergence dynamics that obscure Secondary Opinions; in contrast, our Agency provides transparent access to all Prevalent and Secondary Opinions. Grok Heavy emphasizes performance; in contrast, our Agency also values statistical reliability,  transparency, and pedagogical value.\footnote{Given the different objectives of Grok Heavy and our Agency, any direct comparison of performance is difficult. Our anecdotal tests indicate that Grok Heavy performs as well, sometimes better than, Agency for 2.51 (heat transfer) problems, but performs noticeably --- and somewhat inexplicably --- worse in 2.001 (mechanics) problems.}$\Box$ 

There are also several applications of multi-agent LLM ideas within the mechanical engineering context. In many cases, the emphasis is on heterogeneous architectures, with hierarchical agents for different tasks: examples include MechAgents for elasticity and finite-element modeling \cite{mechagents2023}, and multi-agent systems for robotics \cite{chen2025}. The work reported in \cite{tian2024} combines heterogeneous agents with {\em multiple instances of a particular agent} (or Expert), and is thus quite similar in spirit to our own efforts; however, the context and application of \cite{tian2024} --- LLM-based finite element analysis coordination --- and our own work --- mathematical modeling for Initial Analysis --- is quite different and indeed complementary. All these prior examples of multi-agent systems for mechanical engineering take good advantage of interaction between agents for comparison and resolution --- a concept we adopt in Agency $N$-Plus-$1$ in the form of Agent Compare; in our work, we further propose an underlying quantitative (probabilistic) framework for convergence in number of (Agent Solve) agents.

\medskip

\noindent\textbf{Roadmap of Paper}: We now provide a roadmap for the rest of the paper. In Section \ref{sec:PSPS} we define more precisely what we mean by a Problem Statement and Proposed Problem Solutions; we then introduce, in Section \ref{sec:agency}, Agency $N$-Plus-1; we next describe, in Section \ref{sec:proba}, the underlying probabilistic framework which informs the design of Agency; finally, in Section \ref{sec:futurework}, we discuss future work. Our discussion incorporates examples --- a few of the many examples we have explored --- to justify our claims; the appendices provide a more detailed description of these examples. $\Box$

\section{Problem Statements and Problem Solutions}\label{sec:PSPS}

\subsection{Subject Matter}

In this work we consider two mechanical engineering disciplines, (a) mechanics and (b) heat transfer, as typified in   respective MIT Mechanical Engineering undergraduate subjects (a) 2.001, {\em Mechanics and Materials I} and (b) 2.51, {\em Intermediate Heat (and Mass) Transfer}. However, our approach should be applicable to mechanical engineering analysis more broadly, for example in the disciplines of dynamics and control, fluid mechanics, and thermodynamics. We focus on problems which require the {\em development} --- not just the application --- of appropriate mathematical models.

\subsection{Conceptual Framework}

\subsubsection{Problem Statements} A typical Problem Statement will describe the mechanical engineering analysis problem in natural language and images: the artifact, the environment, and the operating conditions. The Problem Statement must  specify the Quantity of Interest (QoI), or perhaps several QoI; a Problem Statement may or may not include Problem Statement Engineering Context, defined as the way in which the QoI would serve ultimate engineering goals. The Problem Statement should incorporate, either explicitly or implicitly, Problem Statement Parameters, for example related to the artifact geometry and composition,  in a prescribed Problem Statement Parameter Domain.  Finally, the Problem Statement must explicitly or implicitly be inspired by an actual Problem Statement Physical System: the Problem Statement Physical System, represented in appropriate text and images, should correspond to an (admissible) instance of the Problem Statement Parameters; ideally, the Problem Statement may include some experimental Problem Statement Companion Measurements (associated with the Problem Statement Physical System).

We suppose that Problem Statements are aggregated in Problem Classes. In this paper, we will consider two Problem Classes: Problem Class 2.001, which consists of Problem Statements typically encountered in MIT Mechanical Engineering subject 2.001; and Problem Class 2.51, which consists of Problem Statements typically encountered in MIT Mechanical Engineering subject 2.51. We also introduce, for future reference, two Ensembles of Problem Statements: Problem Statement Canon 2.001 (respectively, Problem Statement Canon 2.51) is a large set of 2.001 (respectively, 2.51) problem sets, quizzes, and exams. 

\subsubsection{Problem Solutions}

A typical Proposed Problem Solution will comprise four {\em Parts}; each Part is the result (or output) of a corresponding {\em Stage} of the Proposed Problem Solution Process; each Stage is executed by a Problem Solver  --- currently human. We provide here the four Stages of the Proposed Problem Solution Process:
\begin{itemize}
\item[Part-Stage 1] {\em Data} Completion: Problem Solver provides additional information to ``close'' the Problem Statement such that a Proposed Problem Solution exists. $\Box$
\item[Part-Stage 2] Development of the {\em Mathematical Model}: Problem Solver provides a Mathematical Model which must  yield, for any given admissible Problem Parameters,  the Quantity of Interest (QoI). 
\item[Part-Stage 3] Development  of the {\em Mathematical Solution Procedure}: Problem Solver provides a Solution Script which upon execution yields an exact {\em or approximate} value for the QoI (as defined by the Mathematical Model) in numeric or symbolic form; we denote by $\varepsilon$ the difference between the exact QoI associated to the Mathematical Model and the approximate QoI returned by Solution Script.
\item[Part-Stage 4]   {\em Verification and Validation}: Problem Solver provides arguments for 
\begin{itemize} 
\item[] Verification: confirmation that Solution Script correctly implements the Mathematical Solution Procedure, as well as estimation of the approximation error $\varepsilon$. 
\item[] Validation (or equivalently, Justification): argument that the exact QoI associated with Mathematical Model will (or would) approximate well Problem Statement Physical System experimental measurements for all admissible Problem Statement Parameters. Validation may take the form of {\it a priori} conditions or {\it a posteriori} confirmations --- for example, related to behavior regime --- and will often involve stability or sensitivity analysis.
\end{itemize}
\end{itemize}
In general, the four Stages of the Proposed Problem Solution Process are not entirely sequential: as the Mathematical Model is developed, further Problem Statement Data Completion may be required. Note that there is, of course, no unique way to represent a Proposed Problem Solution. We therefore introduce, for any given Proposed Problem Solution, the associated Equivalence Class under semantic equivalence. In fact, when we refer to a Proposed Problem Solution, we implicitly refer to the  Equivalence Class with which the Proposed Problem Solution is associated.

We introduce two restrictions on the relation between the Problem Statement and any Proposed Problem Solutions. First, the Problem Statement should not be prescriptive as regards any Proposed Problem Solutions: the Problem Statement should make no reference to possible Mathematical Models or Mathematical Solution Procedures. Second, 
Problem Statement Companion Measurements may be invoked only in  Stage 4 (Validation) of the Proposed Problem Solution Process; only measurements {\em in addition to} Problem Statement Companion Measurements --- part of Data Completion --- may serve in the development of the Mathematical Model. 

We also publish our general Expectations for Proposed Problem Solutions. In particular, we ask that, for all admissible Problem Statement Parameters, a Proposed Problem Solution provide a value (or values) for the QoI {\em but also} include a summary of each Part of the Proposed Problem Solution. Of particular importance is Part 4, Verification and Validation, which justifies Part 1, Part 2, and Part 3. We can also include in the Expectations for Proposed Problem Solutions lower-level requirements, such as dimensional homogeneity, a particular system of units, or a citation/attribution convention (say) for Data. These Expectations for Proposed Problem Solutions are intended to guide the Problem Solver and also, ultimately, inform the assessment of any given Proposed Problem Solution. Note that, due to these Expectations of Proposed Problem Solutions, even questions in ostensibly simpler format --- for example, True/False or Multiple Choice --- are only somewhat easier (by virtue of implicit hints) than more standard Freestyle Problem Statements.

We now presume that we are provided with a resource constraint in the form of Allowable Problem Solution Cost (in units of \$): the Problem Solution Cost is defined in terms of the cost associated with any data (and perhaps data acquisition), the cost associated with software fees, the cost associated with computer resources, the cost associated with human effort, and finally any opportunity costs. The human effort would typically be measured by the usual product: effective hourly wage $\times$ Time Worked. The opportunity costs would reflect any implicit losses incurred due to the Elapsed Time required to develop the Proposed Problem Solution. Note both Time Worked and Elapsed Time can be measured either as the average Time over many instances of the Problem Statement Parameters, or more precisely as an initial Time and a marginal Time per Problem Statement Parameter instance.

We do not prejudge the set of possible Mathematical Models. In principle, a Proposed Mathematical Model could be as simple as an interpolant or regression of experimental data or as complicated as a set of nonlinear partial differential equations or, as will often be the case, some combination of the former and the latter. It is important, however, to recall that the Proposed Mathematical Model must address  all Problem Statement Parameters in Problem Statement Parameter Domain:  there is thus an implicit {\em generalizability requirement}. This diversity of potential Proposed Mathematical Models also illustrates the importance of Allowable Problem Solution Cost in rationally accommodating many possible Proposed Mathematical Models; absent the Allowable Problem Solution Cost constraint, the Bernoulli beam, a workhorse of structural analysis, would never be preferred to (say) full three-dimensional elasticity analysis.


\subsection{Initial Analysis}


As already indicated, there will be range of Problem Solutions as we vary Allowable Problem Solution Cost from smaller to larger values. 

Proposed Problem Solutions associated with larger values of Allowable Problem Solution Cost will typically serve in {\em final} design and optimization studies, often with particular reference to both failure and performance. We shall refer to such studies as ``Final Analysis''. 

Proposed Problem Solutions associated with smaller Allowable Problem Solution Cost will serve in many and diverse contexts: as a first estimate of QoI for purposes of (say) {\em conceptual} design, control, or optimization; in various online capacities, for example real-time control; as an inexpensive screen to identify important Problem Statement Parameters for subsequent Final Analysis; for Verification of (often much less transparent) Final Analysis QoI. We shall refer to such studies as ``Initial Analysis''. It should also be clear that the importance of the QoI in Problem Statement Engineering Context will also play a role in the choice of Allowable Problem Solution Cost: if the Problem Statement Engineering Context is not overly sensitive to the QoI, then an Initial Analysis may suffice even as a final analysis.

We now define the limits of Initial Analysis in a more actionable fashion in the form of Initial Analysis Restrictions: the Data Completion should be restricted to readily available archival experimental data (or associated correlations); the Mathematical Model, and any Verification or Validation studies, should incorporate only relatively small algebraic systems, relatively small systems of ordinary differential equations (initial-value or boundary-value),  and a (or a few)  partial differential equations in time and {\em one} spatial variable. The limits on Mathematical Model should suffice to also suitably proscribe expensive and opaque Mathematical Solution Procedures. We emphasize that, in general, the costs of Data Completion can predominate the Problem Solution Costs for higher fidelity analyses; Initial Analysis is as much about elimination of unnecessary data as elimination of unnecessary degrees of freedom.

{\em We consider in our work, and in the remainder of this paper, only Initial Analysis.} Initial Analysis techniques are not new, however the motivations for Initial Analyses, as described above, remain valid --- perhaps even more valid --- within the modern digital computational era. The MIT Mechanical Engineering analysis subjects, particularly in our undergraduate curriculum, focus almost exclusively on Initial Analysis, though more recently with increasingly frequent complementary instances of Final Analyses.

\subsection{An Example}

\begin{figure}[!ht]
\begin{mdframed}
{\bf PROBApple.} We consider the apple shown in Figure 1. The apple sits on a counter in a kitchen. The apple is initially at 280.15K. The surrounding air and kitchen walls are at 296.35K. We wish to predict the average temperature of the apple as a function of time. {\color{lightgray} Elective (slightly prescriptive) instruction: Be as precise as possible in your analysis of the heat transfer coefficient.}
\begin{center} \includegraphics[width = 0.666 \textwidth]{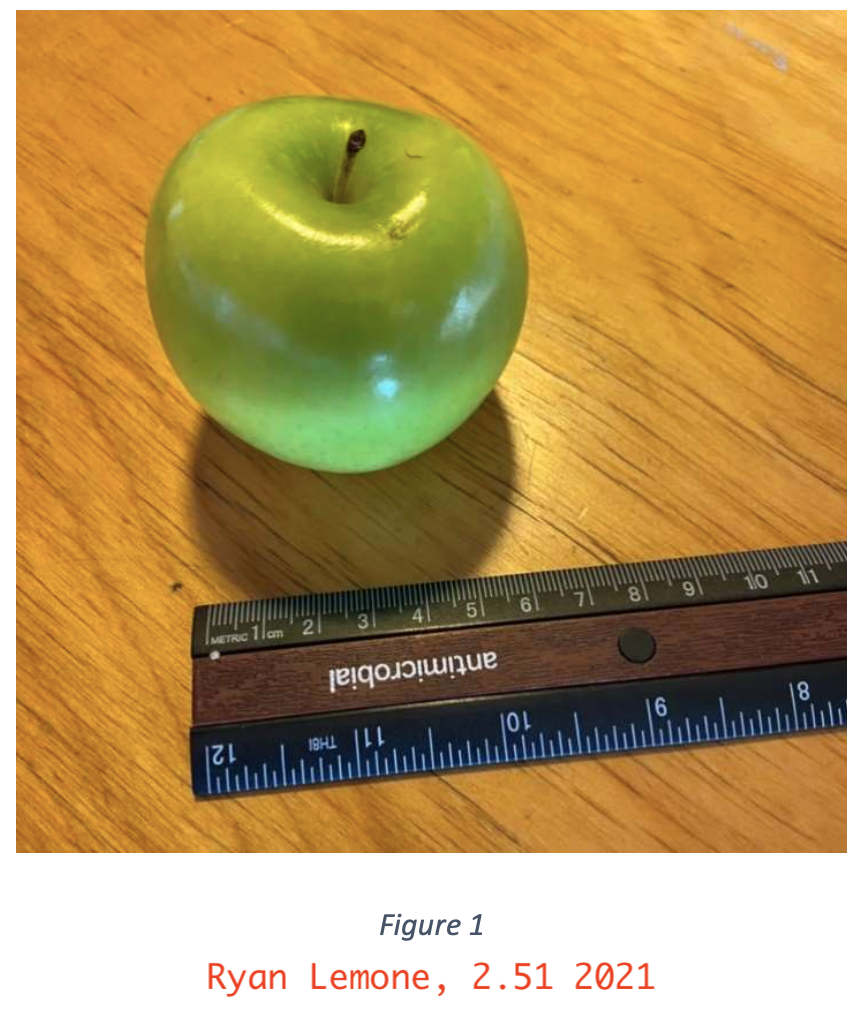}\end{center}
\end{mdframed}
\caption{\label{fig:PROBApple_def} Problem Statement  PROBApple. Note this problem derives from a 2.51 student project (private communication, R Lemone, 2.51 2021).}
\end{figure}

We consider in Figure \ref{fig:PROBApple_def} a Problem Statement associated with MIT Mechanical Engineering subject  2.51, {\em Intermediate Heat and Mass Transfer}: PROBApple. In this example, the Problem Statement Parameters are implicit: the initial temperature of the apple; the  temperature of the ambient; the size of the apple. However, the Problem Statement image explicitly introduces (an instance of) the Problem Statement Physical System. In fact, even Problem Statement Companion Measurements are available (private communication, R  Lemone, 2.51 2021); for brevity, we  omit the Problem Statement Companion Measurements. We choose this ``warming of an apple" Problem Statement because the situation can be well understood without any specialized knowledge; however, similar Problem Statements can be developed for more relevant Problem Statement Engineering Contexts, for example cooling of a 3D Printer Bed (private communication, C Cummings, 2.51 2022) or solar heating of a Camera Aeroshell (private communication, J Benke, 2.51 2023).

Problem Statement Physical System is multiscale --- primarily through the fluid flow and heat transfer, not explicitly visualized in the Problem Statement image --- but nevertheless well represented by a continuum description. Arguably a Final Analysis Mathematical Model is a set of (parametrized) partial differential equations in time and three space dimensions: transient heat conduction in the solid (apple) domain;  the Boussinesq coupled energy and momentum equations and Radiative Transfer Equation in the fluid/air domain. The Problem Solution Cost for Final Analysis would be very substantial as regards both Data Completion --- identification of the detailed geometry and composition of the apple and the counter --- and of course also (implementation and execution of) the Script Solution associated with the Mathematical Solution Procedure --- for example, based on finite element techniques. 

In contrast, and by construction, Initial Analysis Restrictions ensure small Proposed Problem Solution Cost. However, and always, we are not ensured that Initial Analysis will yield sufficiently accurate --- useful in the Problem Statement Engineering Context --- QoI predictions: there are certainly situations in which Initial Analysis is not meaningful. In the case of PROBApple, Initial Analysis can yield (first-order) time constants which are within 35\% of the value deduced from Problem Statement Companion Measurements --- arguably sufficient for purposes of snack planning (or, more generally, food safety purposes). Similar errors are observed for the more relevant Problem Statements associated with cooling of a 3D Printer Bed and heating of a Camera Aeroshell.

\subsection{External Assessment}

We emphasize that our Agency $N$-Plus-$1$, by construction, will not require any external guidance, other than Expectations for Proposed Problem Solutions and Initial Analysis Restrictions, to arrive at a Proposed Problem Solutions and ultimately a Recommended Problem Solution. However, in order to assess the performance of our Agency, and indeed to systematically improve the performance of our Agency, we will require {\em independent} assessment.

Towards that end, let us now assume that we have in hand, for a particular Problem Statement and admissible Problem Statement Parameters, an Initial Analysis Proposed Problem Solution.  Let us further assume that we are equipped with an associated Problem Solution Grading Template: the Problem Solution Grading Template, developed by an expert, is informed by the Expectations for Proposed Problem Solutions and the Initial Analysis Restrictions as applied to our particular Problem Statement. The Problem Solution Grading Template Problem assigns points --- Total Points Available  = 100 --- and subsequently a Grade $G, 0 \le G \le 100,$ equal to the number of points awarded. We say that a Proposed Problem Solution is Correct if $G \ge G_{\text{threshold}}$ (assumed, for simplicity, to be a constant independent of Problem Statement), and Incorrect if $G < G_{\text{threshold}}$. We can of course also consider more nuanced measures of performance, for example categorization by letter grade.

We emphasize that there may be more than one, even several or many, Correct Proposed Problem Solutions (more precisely, many Correct Proposed Problem Solution Equivalent Classes); we assume all these Correct Proposed Problem Solutions are appropriately reflected in the Problem Solution Grading Template. This multiplicity can arise either due to the existence of several viable Mathematical Solution Procedures (for a given Mathematical Model) or
due to the existence of several justifiable Mathematical Models. We expand on the latter: a typical Problem Statement will reflect epistemic uncertainty; different justifiable Data Completions will, in turn, yield different Mathematical Models. Epistemic uncertainty should be expected --- no Problem Statement will anticipate all the information, often not accessible to inspection, required to fully characterize the Problem Statement Physical System; justifiable Data Completion is a crucial Stage of the Proposed Problem Solution Process. Although our simple assessment procedure would award full credit to any Correct Proposed Problem Solution, Initial Analyses that recognize and address epistemic uncertainty and associated multiple Correct Proposed Problem Solutions certainly deserve extra credit.

We note that in the case in which Problem Statement includes Problem Statement Engineering Context, the Problem Statement Engineering Context may be reflected in the Problem Solution Grading Template. Take as an example a Problem Statement which asks for ``{\color{lightgray}a conservative estimate for} the maximum weight of an individual who can be safely supported by a cane of given dimensions and material," where the gray text could be included as a hint or could be omitted (if we wish to fully honor the prescription-free constraint). In this case, without the Problem Statement Engineering Context, there would certainly be epistemic uncertainty related to appropriate boundary conditions for the buckling load calculation; different Data Completions would yield different Correct Proposed Problem Solutions. However, with the Problem Statement Engineering Context, and the explicit or even implicit requirement ``conservative,'' the buckling boundary conditions which yield the smallest associated buckling load are clearly preferred: the epistemic uncertainty is not resolved, but a particular Data Completion is suggested --- with points deducted (in the Problem Solution Grading Template) for other Data Completions.

\subsection{The Generative AI Perspective}

We envision the incorporation of Generative AI, in our case GPT, in mechanical engineering analysis problems as a Co-Problem Solver: GPT would be the first Co-Problem Solver and in particular would execute as ``first analyzer''  all four Stages of  the Proposed Problem Solution Process (described above) to yield a GPT-Proposed Problem Solution; the human would be the second, or consulting, Co-Problem Solver, responsible only for a new Stage in the Proposed Problem Solution Process, 
\begin{itemize}
\item[Part-Stage 5] {\em Confirmation and Acceptance} (or Rejection) of the GPT-Proposed Problem Solution.
\end{itemize}
More generally, we could admit shades of gray, in which Stage 5 is an interaction between GPT and the human, thus affording the opportunity for correction and revision. However, at least in initial embodiment, we prefer a clear division of labor.

Our extended discussion of Problem Statements and Proposed Problem Solutions is, in fact, a prerequisite for any systematic incorporation of Generative AI in mechanical engineering analysis.  First, we must understand what we mean by a Problem Statement and Proposed Problem Solution --- and we must characterize the resources and hence ``type'' of Proposed Problem Solution we seek --- in order to assess the performance of GPT as first Co-Problem Solver. We argue that the challenges for GPT are different for Initial Analysis, our focus here, and Final Analysis (for example, \cite{tian2024}): Initial Analysis places much greater emphasis on ``creative''
mathematical modeling. Second, the advent of Generative AI very substantially changes the calculation of Problem Solution Cost: in principle, the costs associated with human cognition, often the largest component of  Problem Solution Cost in Initial Analysis, are now dramatically reduced --- restricted to new Stage 5 of the Proposed Problem Solution Process. However,  and as already motivated in Section \ref{sec:motivation}, Stage 5 of the Proposed Problem Solution Process presents substantial difficulties --- difficulties we hope to mitigate in our proposed work. 

\section{Agency $N$-Plus-$1$}\label{sec:agency}

\subsection{Description}

In this work, we develop a GPT Agency $N$-Plus-$1$, henceforth Agency, for critical solution of mechanical engineering analysis problems.  Agency here refers to a collection of collaborating GPT Agents. Agency accepts a problem in the form of a natural language/image Problem Statement, as described in detail in Section \ref{sec:PSPS}, and returns a Recommended Problem Solution. We can give Agency an option -model which specifies the underlying GPT model which, through the OpenAI API, and as moderated by the Agency agents (clients), develops the Recommended Problem Solution; absent any -model specification, we assume currently (default) model \texttt{o4-mini, reasoning\_effort=high}.

Our presentation here is informed by relatively extensive application of a Prototype Agency $N$-Plus-$1$ to a (small) Prototype Problem Statement Canon 2.001 and Prototype Problem Statement Canon 2.51. The Problem Statements in Prototype Problem Statement Canon 2.001 and Prototype Problem Statement Canon 2.51 are equipped with associated Problem Solution Grading Templates, in both cases prepared and applied (in the grading context) by an expert. Recall that, for a given Problem Statement, the Problem Solution Grading Template is not known to Agency, and plays no role in Agency deliberations; the Problem Solution Grading Template serves only to independently assess the performance of Agency (for some given -model option) applied to Problem Statement. 

\begin{figure}[!ht]
\begin{center} \includegraphics[width = 0.988 \textwidth]{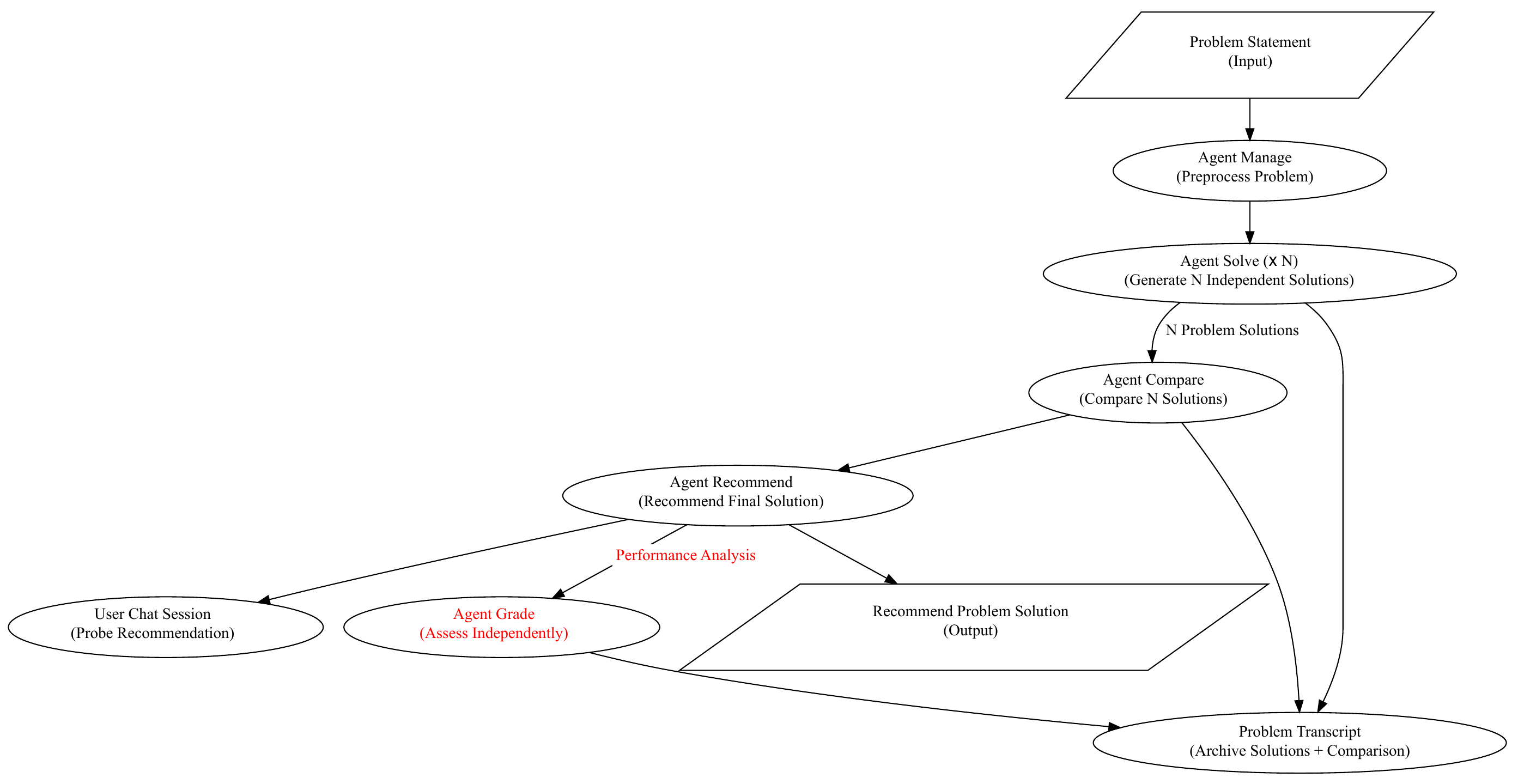}\end{center}

\caption{\label{fig:Flowchart2} Architecture of Agency $N$-Plus-$1$. Note that this figure depicts the ultimate Agency envisioned. The Prototype Agency exercised in this paper is somewhat more primitive: Agent Compare and Agent Recommend are represented together in a single Agent Compare; there is not yet an Agent Grade (see Future Work) or User Chat Session.}
\end{figure}

We summarize the architecture of Agency $N$-Plus-$1$; see Figure \ref{fig:Flowchart2} for an associated flowchart. Prototype Agency $N$-Plus-1 is a somewhat more primitive version of Agency depicted in Figure \ref{fig:Flowchart2}; unless otherwise indicated, Agency shall hereafter refer to Prototype Agency. As already described, Prototype Agency consists of several interacting GPT Agents;  Agent Manage accepts a Problem Statement and performs necessary pre-processing. Agent Manage then invokes $N$ independent instantiations of Agent Solve to obtain $N$ Problem Solution Realizations (a Problem Solution Realization shall specifically refer to an Agent Solve GPT-Proposed Problem Solution). Agent Manage then concatenates the $N$ Problem Solution Realizations into a single file and sends the resulting compilation to Agent Compare to obtain a Recommended Problem Solution along with extensive discussion/comparison of the $N$ Problem Solution Realizations. Agency also composes a complete Problem Transcript of the session, which includes both the $N$ Agent Solve Problem Solution Realizations and the Agent Compare Recommended Problem Solution and associated discussion.  Note that an instantiation of Agent Solve must be treated as random, yielding a Problem Solution Realization; similarly, an instantiation of Agency must be treated as random, yielding an Agency Realization.

Here the name ``$N$-Plus-1" refers to the $N$ instantiations of Agent Solve and the single instantiation of Agent Compare, and inasmuch highlights the two key aspects of Agency $N$-Plus-1: multiple Problem Solution Realizations, which will amplify GPT success probabilities through the mechanism of Predominant Opinion; critical comparative analysis of Problem Solution Realizations, which can complement Predominant or Prevalent Opinion with judiciously interpreted elements of Secondary Opinions. We provide the underlying probabilistic framework in Section \ref{sec:proba}. Note that we can view Agency as homogenous-heterogeneous in agent composition: homogeneous in the $N$ identical instantiations of Agent Solve; heterogeneous in the incorporation of both Agent Solve(s) and Agent Compare. (To our knowledge, Grok Heavy is heterogeneous, without a homogeneous component.)

Each Agent in Agency is provided with Agent Instructions which describe the expected actions and deliverables, as well as perhaps Agent Tools (such as Python environment and specific Python codes). The instructions for all Agents should include the Initial Analysis Restrictions and also Expectations for Proposed Problem Solutions. (In Prototype Agency,  Initial Analysis Restrictions and Expectations for Proposed Problem Solutions are imposed only weakly:  Initial Analysis Restrictions are present by virtue of various limitations in GPT resources (for example the limited Python environment); in future work, we will more strongly enforce, in Agent Instructions, the Initial Analysis Restrictions as well as the Expectations for Proposed Problem Solutions.

As described in Section \ref{sec:motivation}, Agency is intended to substantially improve the performance of ``out-of-the-box'' GPT. We thus require some reference for out-of-the-box GPT -model performance. Towards that end, for any given Problem Statement we shall consider the application of Agency for $N=1$ (and without regard to Agent Compare Recommendation) --- which we shall denote Agent Solve {\it ex situ}.\footnote{In practice, even for $N = 1$, Agent Manage still does perform standard text and image preprocessing, which we implicitly incorporate in the term Agent Solve {\it ex situ}.} Note, however, that Agent Solve, and hence Agent Solve {\it ex situ}, is not quite equivalent to GPT -model ``out-of-the-box'':  Agent Solve already in Prototype Agency is imbued with certain Agent Instructions and Agent Tools --- germane to all Problem Statements in Problem Class --- which  prevent elementary errors and non-substantive distractions: most notably, Agent Solve's Instructions ask GPT to perform any numeric calculations by first writing and then executing a Python script; we thus avoid the nuisance of error-prone GPT ``arithmetic in prose.'' 

The general framework of multiple ``observers,'' in our case multiple instantiations of Agent Solve,  has a rich statistical lineage, indeed dating back to Condorcet's Jury Theorem  \cite{condorcet1785} (see also \cite{meir2022} for recent embodiments). Multiple realizations provides for fault tolerance: given a sufficiently good, but imperfect GPT (default) model, consensus among Agent Solve Problem Solution Realizations is evidence that the Predominant Opinion in fact constitutes a Correct Proposed Problem Solution. We furthermore contend that, even in contexts in which there will be no Predominant Opinion --- for example, Problem Statements with substantial epistemic uncertainty,  Problem Statements in which a key piece of physics is not explicit in the language or images (or well represented in the Pre-Training Set), or Mathematical Models which admit a variety of Mathematical Solution Procedures --- Agent Compare can often provide a solid Recommended Problem Solution; we refer to the apple-warming example of Appendix Section \ref{sec:appExamples} as evidence. In short, a statistical framework transforms the creative aspect of Generative AI from a liability to a strength even in the context of analysis. We discuss the probabilistic framework in greater detail in Section \ref{sec:proba}.

We provide, in Appendix Section \ref{sec:appExamples}, several use cases of Prototype Agency which support the claims and conjectures in this paper. We also include, in Appendix Section \ref{sec:appCompareTranscript}, the Agent Compare part of the transcript associated with an example presented in Appendix Section \ref{sec:appExamples}.


\subsection{Desired Behaviors}

We envision several audiences for Agency $N$-Plus-$1$, which we characterize in terms of the role of the Agency $N$-Plus-$1$ User. Most generally, our primary intended audience is User = Engineering Practitioner, or Practitioner for short. We also introduce the special role of (Agency) Developer. Finally, we  can identify two roles specific to education: User = Student, and User = Instructor. In this regard, it is important to clarify the education aspect of the proposed work. We do {\em not} design Agency as a tutor to help Student develop Proposed Problem Solutions, though if Agency can  serve in this capacity, in particular in the shorter term, all the better. Rather, we design Agency to reduce the role of User --- Practitioner and Student --- from Problem Solver to Co-Problem Solver. The latter role then suggests the crucial education conversion, and associated conversation, which we hope will be engendered by Agency:  How do we educate Student (ultimately, Practitioner) to effectively {\em confirm and accept, or reject} a GPT-Proposed Problem Solution? We discuss below the importance of ``effectively'' in this context.

We enumerate here the key desired behaviors for Agency. Many of these behaviors are already demonstrated, albeit in a limited number of instances, in our Prototype Agency (see  Appendix A, Examples). The future work described in a later section is designed to further improve performance with respect to the metrics introduced below.

\begin{itemize}
\item[B1] With respect to User = Student or User = Practitioner, Agency should (i) save User time, and (ii) spare User drudgery.
\end{itemize}
We note a Behavior B1 is a high-level outcome which incorporates many objectives. For example, as a necessary condition for Behavior B1, Agency must provide Correct Agent Compare Recommended Problem Solutions.
\begin{itemize}
\item[B2] With respect to User = Student or User = Practitioner, Agency should provide information which assists User in the effective assessment of  Agency, and in particular effective assessment of the Agent Compare Recommended Problem Solution. 
\end{itemize}
We pause to emphasize the connection between B1 and B2 via the adjective ``effective'' (alternatively, ``efficient'') in ``effective assessment.'' If User, in order to confirm (or not) the Agent Compare Recommended Problem Solution in Stage 5, Behavior B2, must develop their own Proposed Problem Solution to Problem Statement {\em from scratch and in its entirety}, then Agency does not save User time, Behavior B1, and indeed Agency serves little practical purpose. Agency --- for example, through Agent Compare supporting evidence --- must provide a framework conducive to, and supportive of, User critical judgement.
\begin{itemize}
\item[B3] With respect to User = Developer, Agency should permit Developer to readily monitor GPT performance 
as new GPT models are released. Developer is then equipped to update GPT default model in an informed fashion.

\item[B4] With respect to User = Instructor, Agency should facilitate inexpensive Student access to best GPT models at scale.
\end{itemize}
We also include several collateral education-specific  behaviors:
\begin{itemize}

\item[B5] With respect to User = Instructor, Agency should assist Instructor in the development of Problem Statements which (i) are free of unproductive ambiguity and convention-specific knowledge, and furthermore (ii) promote clear assessment of desired Student outcomes.

\end{itemize}
We note that many of these behaviors also require complementary infrastructure efforts, for example to provide easy access to Agency.

\section{A Probabilistic Framework} \label{sec:proba}

\subsection{Formulation}

We provide here a simple probabilistic framework with which we can understand and anticipate the performance of Agency. In fact, this probabilistic framework can ultimately be (but is not yet) explicitly incorporated into Agent Compare.

We first consider a given Problem Statement. We suppose that we are provided with $N$ Agent Solve Problem Solution Realizations which, for purposes of convenience and simple exposition, we suppose can be categorized in two sets, Set 1 and Set 2, where Set 1 contains $M_1$ Problem Solution Realizations semantically equivalent to a Proposed Problem Solution 1 (alternatively, in Proposed Problem Solution 1 Equivalence Class), and Set 2  contains $M_2$ Problem Solution Realizations semantically equivalent to a Proposed Problem Solution 2 (alternatively, in Proposed Problem Solution 2 Equivalence Class). We further suppose that  Proposed Problem Solution $I$, for $I \in \{1,2\}$, is a Correct Proposed Problem Solution, that Proposed Problem Solution 1 is semantically different from Proposed Problem Solution 2, that there is only a single Correct Proposed Problem Solution (Equivalence Class) to Problem Statement, and hence that Proposed Problem Solution $I' \equiv 3 - I$ is an Incorrect Proposed Problem Solution. (Note that the final conclusion relies on our assumption that there is a single Correct Problem Solution Equivalence Class for our given Problem Statement; we consider the implications of two semantically different Correct Proposed Problem Solutions shortly.) Finally, we choose $M_1 \ge M_2$ (ties resolved by a fixed rule); also, by construction, $M_1 + M_2 = N$. We denote, based on our convention, Proposed Problem Solution 1 as the Predominant Equivalence Class, and Proposed Problem Solution 2  as the Secondary Equivalence Class. Note that in our current restricted discussion, Proposed Problem Solution Equivalence Class 1 and Proposed Problem Solution Equivalence Class 2 are fixed, however each Agency Realization will yield different $I$ (and $M_1$, $M_2$).

Next let us introduce a Bernoulli random variable $B$ associated to our given Problem Statement: $B =1$, Pr$(B=1)= p$ --- Agent Solve -model {\it ex situ} provides a Correct Problem Solution Realization; $B = 0$, Pr$(B=0) = 1-p$, Agent Solve -model {\it ex situ} provides an Incorrect  Problem Solution Realization. Recall that Correct (respectively, Incorrect) are dictated by grade $G\ge G_{\text{threshold}}$ (respectively, $G < G_{\text{threshold}}$) as decided by application of Problem Solution Grading Template by an expert. In our current context, we can identify Correct with Proposed Problem Solution $I$, and Incorrect with Proposed Problem Solution $I'$. We can then readily evaluate the probability (implicitly conditional on $N$)
\begin{align}
\Pr(\text{Proposed Problem Solution 1 is Correct}\,|\, M_1)\hspace{1in}
\nonumber \\[.2em] = \; \frac{p^{M_1}(1-p)^{N-{M_1}}}
      {p^{M_1}(1-p)^{N-M_1}+(1-p)^{M_1}p^{N-M_1}}  \nonumber \\[.2em] = \;\dfrac{1}{1 + \left(\dfrac{1-p}{p}\right)^{2M_1 - N}}\;\quad\quad,\hspace{.728in} \nonumber \\[.2em]
      = \dfrac{1}{1 + \left(\dfrac{1-p}{p}\right)^{(2 \rho_0-1)N}}\;\quad\quad,\hspace{.728in}  \label{eq:Con}
\end{align}
where we take advantage of $\binom{N}{M} = \binom{N}{N-M}$ to simplify our expressions; here $\rho_0 = M_1/N$. Note that the event $$\{\text{Proposed Problem Solution 1 is Correct}\,|\, M_1\}, $$
equivalent to event $\{ I = 1\,|\, M_1\}$,  is indeed a random event, since $I$ is a random variable; note the sample space here is the set of all possible Agency outcomes --- equivalent to all possible outcomes of $N$ instantiations of Agent Solve.  (In the interest of succinctness, we let context, rather than notation, distinguish random variables from associated random variates.) Note we interpret Equation \eqref{eq:Con} as a frequentist conditional probability, not a Bayesian posterior.

We have not yet taken advantage of the significance of $M_1$, and in particular $M_1 \ge M_2$. In fact, it can be readily shown that, for $p > 1/2$, and for (or, by assumption, {\em since}) $M_1 \ge M_2$, then (from Equation \eqref{eq:Con}), $\Pr(\text{Proposed Problem Solution 1 is Correct}\,|\, M_1) > 1/2$, and hence the Predominant Equivalence Class is more likely to represent the Correct Problem Solution Realization than the Secondary Equivalence Class. Furthermore, for $1 - p$ small and $M_1/N \equiv \rho_0 > 1/2$, the expression on the right-hand side of Equation \eqref{eq:Con} approaches 1 rapidly (for fixed $\rho_0$) as $N$ increases. We provide quantitative examples shortly. Needless to say, the advantage of Predominant Opinion is that we can identify a Correct Proposed Problem Solution without reference to, or application of, any Problem Solution Grading Template.

Note also that we can bootstrap an estimate for $p$ (in the Predominant Regime), $\hat{p}$, as
\begin{align}
\hat{p} \equiv M_1/N \;;  \label{eq:bstrap}
\end{align}
if $p > 1/2$, and in particular if $1-p$ is small, $\hat{p}$ should provide a good estimate for $p$. Here we consider  probability $p$ for a particular Problem Statement, hence $p$[Problem Statement]. However, Equation \eqref{eq:bstrap} can be misleading if $p \not > 1/2$, and hence some prior information would be useful.

Towards that end, we can also consider some metric for Agency performance for an Ensemble, $\varpi$[Ensemble], for example $\varpi$[2.51 Problem Statement Canon]; we might take $\varpi$ as the average of $p$[Problem Statement] over Problem Statements in the Ensemble. Note that the calculation of $\varpi$[Ensemble] is facilitated in our (Prototype) Agency system by a simple MultiProblemSolve \texttt{bash} script which invokes Agency for some given $N$ for a list of Problem Statements. Since $\varpi$[Ensemble] can be calculated {\it a priori}, $1 - \varpi$[Ensemble] small can lend further credence to Predominant Opinion and also the accuracy of our {\it posteriori} estimate $\hat{p}$ for given Problem Statement. As indicated earlier, our interest, and the relevance of the proposed work to mechanical engineering, is Problem Class 2.001 and Problem Class 2.51. We find, for the GPT default model, and our admittedly small Prototype Problem Statement Canon 2.001 and Prototype Problem Statement Canon 2.51, that $\varpi$ is indeed close to $1$, and in any event considerably closer to $1$ than to $1/2$. Note this would not be the case for earlier GPT models in particular of the non-reasoning variety (or even for GPT model \texttt{o4-mini}, \texttt{reasoning\_effort=low}); alternatively stated, Agency is only now, in 2025, viable.

We now consider the incorporation of this probability framework into Agency. Towards that end, we must first make an Agent Compare Discernment Hypothesis: Agent Compare can categorize Proposed Problem Solutions in associated Equivalence Classes. Already in Prototype Agency we have good evidence, see Appendix B, that (for our GPT default model) the Agent Compare Discernment Hypothesis is indeed satisfied. We would then incorporate in Agent Compare Instructions the essential aspects of the probabilistic framework developed here with particular reference to Prevalent Opinion. Alternatively, and as currently implemented in Prototype Agency, we can leave the process which translates the Problem Solution Realizations into a Recommended Problem Solution to the discretion of Agent Compare and, only implicitly, notions of Prevalent Opinion represented in the Pre-Training Set; as we describe in Appendix B, this un-instructed approach actually performs very well, and also is arguably less brittle as the assumptions of our probabilistic framework are relaxed.

In that regard, in the remainder of this section, we relax our suppositions: the Problem Solution Realizations may belong to more than two Equivalence Classes and furthermore there may be multiple Correct Problem Solutions (any of which may be potentially represented in the set of $N$ Problem Solution Realizations). We thus generalize our notation: we introduce $p_1, p_2, \ldots, p_K$ for $K$ (distinct) Correct Problem Solution Equivalence Classes (satisfying the Initial Analysis Restrictions), and then introduce $p_{\max} \equiv \max_{k \in \{1,\ldots,K\}} p_k, p_{\min} = \min_{k \in \{1,\ldots,K\}} p_k$, and $p_{\text{tot}} = \sum_{k=1}^K p_k$; note that $p_{\max} = p_{\min} = p_{\text{tot}} = p$ for the case $K = 1$. We further assume that all Incorrect Proposed Problem Solution Equivalence Classes are characterized (in aggregate) by probability $p^* = 1 - p_{\text{tot}}$.

\subsection{Case A: Unique Correct Solution {\em and} $p_{\max} > 1/2$} 

We first consider the case of a unique Correct Problem Solution Equivalence Class with probability $p_{\max} > 1/2$. In this situation we recover the clean Condorcet-style theory: the Predominant Equivalence Class is, with high probability, Correct: \ $\Pr(I=1\,|\,M_1) > 1/2$ and rapidly approaches $1$ with increasing $N$. 

For illustration, let $p_{\max} = 0.8$, $N = 8$, and $M_1 = 6$. Then, from Equation~\eqref{eq:Con}, we obtain 
\[
\Pr(I=1 \mid M_1=6) = 0.9961, 
\qquad 
\Pr(I=2 \mid M_2=2) = 0.0039 .
\]
The interpretation is straightforward: If the GPT (default) model performs well for the Problem Statement ($p_{\max}$ close to 1), then {\em agreement} among $M_1$ Problem Solution Realizations in the Predominant Equivalence Class is a strong indication that the Predominant Opinion is a Correct Proposed Problem Solution. In actual practice, under our Agent Compare Discernment Hypothesis, Agency $N$-Plus-$1$ will be able to identify this Predominant Equivalence Class. Once the Predominant Opinion is revealed, we may then bootstrap our estimate for $p_{\max}$ as $\hat{p} = M_1/N$. Note that even for $p_{\max} = 0.8$, $N = 4$, $M_1 = 3$, Equation~\eqref{eq:Con} yields $\Pr(I=1 \mid M_1=3) = 0.9412$; we do not require $N$ large. 

We find that Agent Compare Recommendations, even without explicit reference to the underlying inference framework, will largely follow the Predominant Opinion. However, since Agent Compare has access to both Part 2 and Part 3 of the Agent Solve Problem Solution Realizations, the Agent Compare Recommended Problem Solution may be largely dictated by the Part 2  but nevertheless nuanced by Part 3. In particular, and as we would expect within the Generative AI context, Problem Solution Realizations which arrive at the same conclusion may nevertheless differ in the (say) Mathematical Solution Procedure: this diversity of perspectives is very beneficial to assessment (Stage 5), and of course also serves as an important pedagogical function. Agent Compare may even comment positively on the Mathematical Solution Procedure associated with a Problem Solution Realization which is not Predominant. In general, Agency $N$-Plus-$1$ is intended to transform the irreproducibility of Generative AI (in analysis contexts) from a disadvantage to an advantage. 

We note also in this context the role of Agent Compare not only as critical analyst but also as summarizer. In practice we would typically not choose $N$ too large, but even for $N = 3$ it would be a time-consuming task for User to review, interpret, and compare $N$ Agent Solve Problem Solution Realizations. And, of course, it is much easier for User to confirm an error identified by Agent Compare in an Agent Solve Problem Solution Realization than for User to develop ``by hand'' a complete Proposed Problem Solution. In this sense, Agency does indeed provide for {\em effective} assessment by User in Stage 5. 

The situation (not just $p_{\max} > 1/2$ but) $p_{\max} \approx 1$ is especially favorable. We can predispose our analyses to fall within this category: we precompute $\varpi$[Ensemble] for a best currently available GPT model; if $\varpi$ is not sufficiently close to 1, we would wait to deploy Agency pending the appearance of an improved GPT model. It is certainly the case that no GPT model --- no Agency --- is better than a very poor GPT model. Conversely, we emphasize that even for $\varpi$ close to $1$, and indeed even $p_{\max}$ close to $1$, Agent Solve {\it ex situ} will, with finite probability, return a Problem Solution Realization which is Incorrect;  hence the importance of multiple Realizations and related inference. 

We illustrate the interplay between GPT model, Agency, $\varpi$, and our MultiProblemSolve Agency utility through a concrete event. OpenAI recently released, after we had largely completed this paper, GPT-5. We were able to easily invoke MultiProblemSolve to deduce that $\varpi$ for GPT-5 (with \texttt{reasoning\_effort=high}) is indeed larger than for $\varpi$ for \texttt{o4-mini-high} with \texttt{reasoning\_effort=high}; GPT-5 with \texttt{reasoning\_effort=high} will now be the Agency GPT default model in future work. How should we adapt to this upgrade in GPT model performance? If we retain the same Problem Canon, we could now consider $N$ smaller than in the past, given the Condorcet result --- {\em though we would still take (say) at least $N=3$} since (i) as emphasized above, failure is possible even for $p_{\max}$ close to unity\footnote{We provide an example of the continued merit of Agency even within a further improved GPT model. We consider a very simple --- first week of 2.001 --- Problem Statement related to elongation of a bar of variable cross section under uniaxial tension. We applied Agency with model GPT-5 (with \texttt{reasoning\_effort=high}) for $N = 3$: only two of the Agent Solve Problem Solution Realizations were Correct --- the third suffered a sign error; Agent Compare chose the Predominant Opinion and furthermore identified the error in the Secondary Opinion.}, and (ii) diversity in Problem Solution Realizations, in both Mathematical Model and Mathematical Solution Procedure, is beneficial to assessment, and (iii) there may still be particular members of the Problem Class for which $p_{\max}$ is relatively low. Alternatively, we can respond to the upgrade in GPT model performance by considering a larger and more difficult Problem Canon, in some sense ``preserving $\varpi$."

\subsection{Case B: Multiple Correct Solutions ($p_{\max} \le 1/2$) {\em and} $p_{\min} > p^*$} \label{sec:favcase}

We next consider the case where there is no single dominant Correct Problem Solution Equivalence Class, but instead several Correct Problem Solution Equivalence Classes, each characterized by probability greater than the Incorrect Equivalent Classes probability $p^*$. In this regime, $p_{\text{tot}}$ is close to unity,  but $p_{\max} \le 1/2$; we should not, in general, expect a Predominant Opinion. (In fact, the arguments of this section are also relevant, if $N$ is modest, to the case in which $p_{\max} > 1/2$ but with a close competitor, say $K=2$ and $p_1 = .51$, $p_2 = 0.48$.)  In such cases, we would hope that Agent Compare will recognize and report the existence of more than one Correct Proposed Solution --- the Correct Recommendation is arguably a set, not a singleton.

{\em Epistemic Uncertainty}. This regime is precisely the mathematical reflection of epistemic uncertainty in engineering analysis: the epistemic uncertainty gives rise to different Data Completions and ultimately different Correct Problem Solutions --- a multi-modal distribution. 
Epistemic uncertainty is very common --- What are the properties of a composite? How, precisely, is the beam attached to the support? How well are two material layers thermally mated? How rough is a pipe wall? User would recognize in different Problem Solution Realizations the influence of epistemic uncertainty and thus the potential validity of different Problem Solution Realizations (hence multiple Correct Problem Solutions): no single choice is the correct Recommendation; the associated uncertainty must be carried downstream in subsequent applications of the QoI. At present, (Prototype) Agent Compare is a bit too rigid, perhaps because the present Agent Compare Instructions request {\em a} Recommendation; the Recommendation language should be softened to recognize and incorporate epistemic uncertainty and hence several possible Recommended Problem Solutions. $\Box$

\subsection{Case C: Unfavorable ($p_{\max} \le p^*$)} \label{sec:unfavcase}

Finally, we consider the case in which any Correct Problem Solution Equivalence Class has probability at most $p^*$. In this unfavorable regime, Incorrect Proposed Problem Solution Equivalence Classes may be just as frequent (or more frequent) than Correct Problem Solution Equivalent Classes. By the same reasoning that led to positive conclusions in Case A, now we might expect Agent Compare to Recommend an Incorrect Proposed Problem Solution  --- so that User (as ultimate arbiter, in Stage 5), presented with a preponderance of ``evidence,'' could well accept the Recommendation. However, although Agent Compare --- and User --- will certainly recognize the Prevalent Opinion as a factor, other factors can also play a role in the assessment, ultimate decision, and implications. 

{\em Missing Physics}. In some cases, effective outlier analysis can save the day. A fairly common error, committed by humans and also by GPT, is to simply ``leave out'' some piece of physics which, in fact, plays an important role. The physics may not be apparent in the wording of the problem or easily visible in the associated images. It is also possible that common omissions are the result of unqualified statements in the literature and textbooks, then propagated into GPT Pre-Training datasets. For example, ``thermal radiation is only important at higher temperatures'' is part of the heat transfer lore, but the statement is not true if natural convection is the primary non-radiation mechanism for heat transfer between a body and the environment. Similarly, second moments of area are often presented as scalars without discussion of the more general tensor relation from which the particular (context-dependent) relevant scalar is deduced. 

We might expect that, for a Problem Statement which excites such misinformation, $p_{\max}$ (and hence also $p_{\text{tot}}$) could be quite small. However, comparison of two Agent Solve Problem Solution Realizations with respectively missing physics and all necessary physics will typically elicit an ``Aha'' response --- immediate recognition of the omission-cum-error --- such that a Secondary Opinion will readily be Recommended/accepted, by Agent Compare and also by User. $\Box$

{\em Scattered Problem Solution Realizations}. Some Problem Statements are difficult for students and also GPT simply because there might be many interacting parts within the description --- in words or images --- either of the artifact in question or of the associated environmental and operational conditions. In such situations, we anticipate that the Problem Solution Realizations will not conform to any obvious pattern or notable predominance. In such a situation, User (and, we anticipate, Agent Compare --- but we do not yet have much evidence) should sense that considerable additional analysis must be undertaken before any Recommendation is proposed or accepted. In this case, Agency might well be inconclusive, but at least in a fail-safe fashion. 

However, scattered Problem Solution Realizations can also be transformed by Agent Compare into a Correct Proposed Problem Solution (Recommendation). We have anecdotal evidence that, even in the case of (say) 1 Correct Solution Realization and $N$ - 1 Incorrect Solution Realizations (all from different Problem Solution Equivalent Classes), Agent Compare may well identify and Recommend the Correct Problem Solution Realization --- consistent with the adage that recollection (by Agent Solve) is more difficult than recognition (by Agent Compare), as already introduced in the context of missing physics. Thus there is merit in $N$ realizations even for $p_{\max}$ small --- justified now by recognition rather than Prevalent Opinion; furthermore, as $N$ increases, the probability of no Correct Solution Realization decreases, thus favoring an Agent Compare ``discovery.''  (Note that the evidence cited here was in fact obtained from first tests of {\em non}-Prototype Agency, armed with both an Agent Compare and an Agent Recommend.) $\Box$

{\em Incorrect Prevalent Opinion}. It is also the case that certain information is rather uniformly and unreservedly absent or Incorrect within the literature, and presumably also in Pre-Training datasets, such that, for related Problem Statements, $p_{\max}$ and $p_{\text{tot}}$ will be very small; hence, for moderate $N$, we would not expect to observe the missing physics in {\em any} Problem Solution Realizations. In this case User, and Agent Compare, may well return an Incorrect (and unqualified) Recommendation --- exactly what we wish to avoid. In this case, our only recourse is to intervene through the mechanism of Agent Solve Instructions. We relegate the design of these ``interventions'' to future work. 

\section{Future Work}\label{sec:futurework}

We enumerate here a few of the more substantive elements for Agency development:
\begin{enumerate}
\item Expansion of Problem Statement format and processing to include explicit Problem Parameters and Problem Parameter Domain. Note that Problem Parameter values can serve in two important capacities: for numerical evaluation of QoI; for symbolic evaluation of QoI but with due regard for ``regime'' as dictated by specific Problem Parameter values.
\item Development of a new Agent Recommend such that Agent Compare focuses only on comparison and discussion and then Agent Recommend focuses only on Recommendation --- perhaps developing a composite of several Agent Solve Realizations. Furthermore, the Agent Compare and Agent Recommend Instructions should be expanded to more explicitly include the considerations of Section \ref{sec:favcase} and Section \ref{sec:unfavcase}.  
\item Incorporation into Agent Solve (and also Agent Compare and Agent Recommend) Instructions an explicit statement of Initial Analysis Restrictions which must be honored as well as Expectations for Proposed Problem Solutions.
\item Implementation of a post-Agent Recommend chat session with User. In this chat session, User can probe Agent Recommendation to better understand the Recommended Problem Solution, and potentially reconcile differences between User and Agent Recommend.
\item Development of an Agent Grade which applies Problem Statement Grading Template to evaluate grade $G$ for a Recommended Problem Solution. Note Agent Grade will be invoked separately from the rest of Agency and the results will be shared only with User, not other Agency Agents: Agent Grade only serves for (non-operational) independent assessment.
\item Study of Problem Statement improvements, with the goal to reduce convention-dependent answers, to eliminate unintended/non-constructive ambiguity, and to promote better assessment  of important Student outcomes (for example, through routine incorporation of some numeric component). In this context, Agency serves as a surrogate student.
\item Enhancement of the underlying statistical inference framework, in particular to consider multi-part questions and outlier analysis. We may also consider a more actionable approach to the probabilistic interpretation --- for example a hypothesis-testing framework --- which could thus assist in (but not dictate) Agent Recommend's decision.
\end{enumerate}
In addition to these major considerations, quality-of-life improvements should also be considered --- in anticipation of deployment more broadly.


\begin{appendices}

\section{Examples} \label{sec:appExamples}

In this appendix we present some examples based on our Prototype Agency $N$-Plus-$1$. Except in one instance (noted) we choose GPT default model. 


In most of our examples we choose $N = 10$ realizations, primarily for illustrative purposes, and to emphasize the probabilistic nature of GPT response. In actual practice we would, in general, choose $N$ smaller than 10, even $N = 2$ or perhaps $N=3$ (to break ties), in order to reduce both cost and also response time. 

There are many Problem Statements, both in Problem Class 2.001 and Problem Class 2.51, many quite difficult, for which Agent Solve {\it ex situ} performs flawlessly. However, the critical discussion point is not the many Problem Statements for which Agent Solve {\it ex situ} performs remarkably and uniformly well, but rather the Problem Statements for which Agency might err --- probability $p$ distinctly less than $1$. We thus  emphasize in our examples here cases in which Agent Solve {\it ex situ} is not perfect.

For all the examples discussed here, and many more, we have complete Agency Problem Transcripts.

\subsection{Example PROBNatlConvPlate}

\begin{figure}[!ht]
\centering
\includegraphics[width = 1.068 \textwidth]{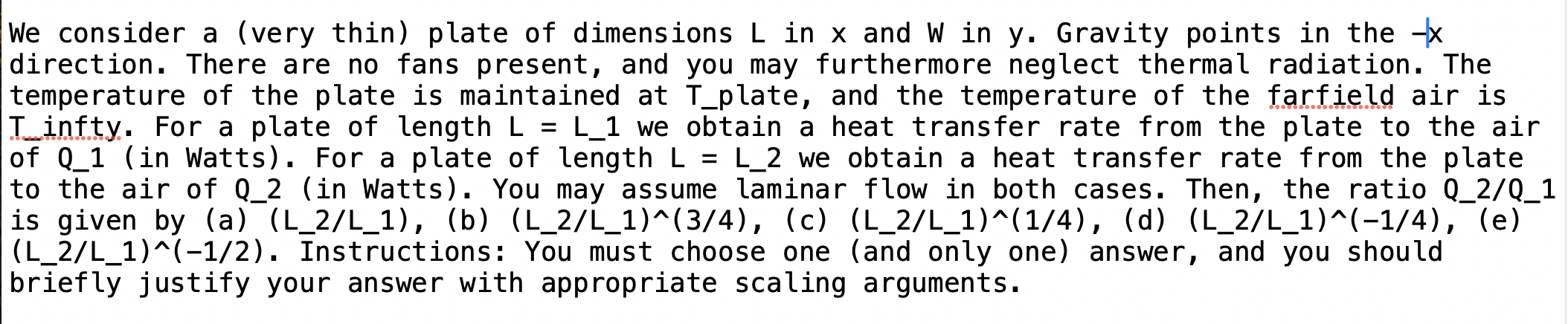}
\caption{\label{fig:PROBnc_def} Problem Statement (.png file) for PROBNatlConvPlateMP.}
\end{figure}

The .png file with Problem Statement provided to Agency is presented in Figure \ref{fig:PROBnc_def}. The question is multiple choice: there is one Correct answer, four Incorrect answers; some of the Incorrect answers are seductive ``traps'' which reflect common student errors in the context of convection scaling analyses more generally. Note a Correct Proposed Solution must include the Correct (multiple-choice) answer but also the supporting material associated with Stages 1-4 of the Proposed Solution Process.

We first consider (default) model \texttt{o4-mini, reasoning\_effort=high}. We choose $N=10$. All 10 instantiations of Agent Solve return the same answer, multiple-choice option (b) and associated justification. As discussed earlier, 10 realizations which return the same answer, here answer (b),  constitutes overwhelming statistical evidence --- under our assumption that $p$ is close to 1 ---  that answer (b) is the Correct answer. And indeed, answer (b) is the Correct answer. In this instance, there is very little that Agent Compare can contribute to the conversation, even given GPT's chatty nature.

Next, to make the example richer, we consider Model \texttt{o4-mini, reasoning\_effort=low}. We again choose $N=10$. Now only 9 instantiations of Agent Solve return the identical  (and Correct) Problem Solution Realization (b); one Problem Solution Realization, Realization 4,  returns Incorrect answer (e). Again, 9 of 10 Realizations which yield the same answer constitutes overwhelming statistical evidence that answer (b) is the Correct answer. Agent Compare of course Recommends answer (b), and, in this sense, Agency with model \texttt{o4-mini, reasoning\_effort=low} is successful.

However, it is instructive to further analyze the errant realization. To arrive at the Incorrect answer (e), Agent Solve in fact makes two mistakes: Mistake 1, Agent Solve applies the scaling result, albeit correctly, for forced convection, not natural convection, to obtain $Q_2/Q_1 = (L_2/L_1)^{1/2}$; Mistake 2, Agent Solve misidentifies $(L_2/L_1)^{1/2}$ with answer (e), though answer (e) in fact corresponds to $(L_2/L_1)^{-1/2}$ (which is the scaling for the heat transfer coefficient, not the heat transfer rate). We might conclude that Agent Solve Realization 4 is quite confident in answer (e) despite implicit clues to the contrary. As already indicated, in this model \texttt{o4-mini, reasoning\_effort=low} case, Agent Compare Recommends answer (b), but Agent Compare also explains where Realization 4 went wrong: ``Discard Solution [Realization] 4 as it applies the wrong physics (forced rather than free convection).''; Agent Compare does not deign to comment on (Rookie) Mistake 2.

\subsection{Example PROBPinnedAssembly}

\begin{figure}[!ht]
\centering
\includegraphics[width = 1.068 \textwidth]{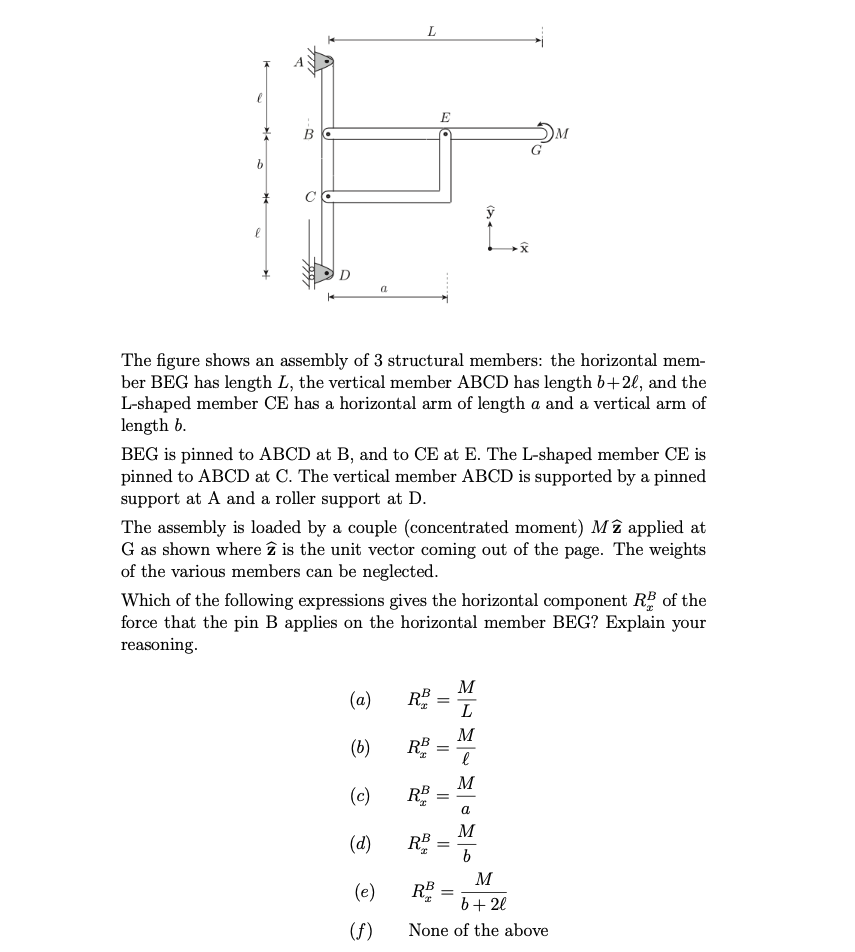}
\caption{\label{fig:PROB3mem_def} Problem Statement (.png file) for PROBStructure\_3Members.}
\end{figure}

The .png file with Problem Statement provided to Agency is presented in Figure \ref{fig:PROB3mem_def}. Note that proper interpretation of graphical information, with respect to both topology and geometry, is crucial. The question is multiple choice: there is one correct answer, four incorrect answers, and also a ``none-of-the-above'' option; the incorrect answers are seductive ``traps'' which reflect common student errors in particular related to the choice of a critical moment arm. We consider default model \texttt{o4-mini, reasoning\_effort=high}.

We choose $N = 10$. Only 8 realizations of Agent Solve return the same answer, answer (d); two Problem Solution Realizations, Realization 4 and Realization 8, return answer (e), ``none of the above.'' Once again, there is very strong statistical evidence that the prevalent answer (d) is the Correct answer. And indeed, answer (d) is the Correct answer. Predictably, Agent Compare recommends answer (d); Agent Compare also identifies the critical error in Realization 4 and Realization 8, and furthermore criticizes Realization 9 for obtaining the Correct answer but based on faulty reasoning. We include the Agent Compare part of the Problem Transcript (verbatim) as Appendix B.

We might further deduce, for our estimate of $p$ for this particular Problem Statement, $\hat{p} = 0.8$. We can plausibly interpret problem statement difficulty as inversely proportional to $p$ (or $\hat{p}$). It is clear from our results, even based on a small sample, that PROBPinnedAssembly (of this section) is more difficult than PROBNatlConv (of the previous section).

\subsection{Cameos}

\subsubsection{PROBApple}

In PROBApple we present a photograph of an apple (next to a ruler), provide appropriate operational and environmental conditions (temperatures of the refrigerator and kitchen), and ask for the average temperature of the apple as a function of time. The reader is referred to Figure \ref{fig:PROBApple_def}. We consider only GPT default model and furthermore $N = 2$. 

In Problem Solution Realization 1, Agent Solve performs very well, but {\em misses} a critical piece of physics --- the thermal radiation contribution to the heat transfer coefficient --- with considerable detriment to the prediction; in Problem Solution Realization 2, Agent Solve again performs very well, but now {\em includes} thermal radiation --- and obtains a much better prediction (for example, compared to experiment).\footnote{Problem Solution Realization 1 and Problem Solution Realization 2 also differ in the method adopted to solve (approximately) the heat equation. Diversity in solution approaches is an advantage of Agency $N$-Plus-$1$, but not the main point in the current example.} Agent Compare indicates ``Strongest recommendation: adopt Solution [Realization] 2.", but in fact Agent Compare is not sufficiently critical of Realization 1, perhaps because the solution approach of Realization 1 is indeed more transparent.

We have already discussed ``missing physics'' above:  If several Agent Solve instantiations yield very similar Problem Solution Realizations, all with the same missing physics, but a single realization yields a different answer now incorporating and justifying the (missing) physics, Agent Compare will most likely side with the Secondary Opinion --- ``missing physics'' is typically obvious once revealed. Once again, multiple Problem Solution Realization and then subsequent  comparison can provide much better assessment than a single Problem Solution Realization. 

\subsubsection{PROBRuler}

\begin{figure}[!ht]
\centering
\includegraphics[width = 1.0618 \textwidth]{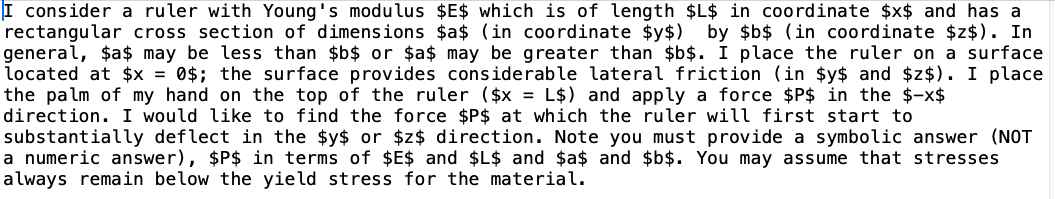}
\caption{\label{fig:PROBRuler_def} Problem Statement (.png file) for PROBRuler.}
\end{figure}

The .png file with Problem Statement provided to Agency is presented in Figure \ref{fig:PROBRuler_def}. We consider only GPT default model and furthermore $N = 2$. 

There are two difficulties presented in this Problem Statement: the first difficulty, technical, is the proper choice of the {\em minimum} second moment of area --- Agent Solve is not fooled; the second difficulty, conceptual, is the choice of boundary conditions on the ruler. In fact, from the word description, the appropriate boundary conditions are not entirely clear, and in any event not unambiguously prescribed. This lack of clarity is an example of epistemic uncertainty --- very common in mechanical engineering analysis --- in which the Problem Statement (indeed, available information) does not include sufficient detail to precisely specify the underlying mathematical model. 

In this example, Problem Solution Realization 1 and Problem Solution Realization 2 choose different boundary conditions (Data Completions), respectively  pinned-pinned and pinned-free. Agent Compare prefers Problem Solution Realization 2, but there is no single Correct answer. This is an example in which multiple Realizations are important not to define a consensus but rather to identify epistemic uncertainty and hence the existence of several possible Correct Problem Solutions. In this context, and indeed more generally with multiple Problem Solution Realizations (and perhaps multiple Mathematical Solution Procedures), Agency $N$-Plus-$1$ embraces --- and takes advantage of --- the  probabilistic and hence creative nature of GPT predictions.

\section{PROBPinnedAssembly:\\ Agent Compare Transcript} \label{sec:appCompareTranscript}

The transcript below was produced {\em entirely} by Agent Compare and is reproduced here verbatim.



\section*{Supplementary material}

=== Comparison ===
All ten write-ups ultimately fell into three camps:

1. Correct, systematic two-member splits that arrive at
$$R_x^B = M/b$$
(i.e. choice (d)). These are Solutions 1, 2, 3, 5, 6, 7, and 10.

2. Attempts that carried the couple's lever arm as L instead of a (or otherwise mis-identified the two-force member), leading to a "None of the above" or to $R_x^B \propto a/(b\,L)$. Those are Solutions 4 and 8, both algebraically inconsistent with the geometry.

3. One very short argument (Solution 9) that collapsed the two pins at B and C into one horizontal-force balance and got the right number, but which glossed over the fact that CE is not a simple horizontal two-force link. Its logic is incomplete even though its final formula happens to be M/b.

Here is a high-level comparison of the best (correct) methods:

Solution 1
- Break into FBD of BEG and FBD of CE.
- Full $\Sigma F_x$, $\Sigma F_y$, $\Sigma M$ at B on BEG ⇒ $R_e^y = M/a$.
- Full $\Sigma M$ at C on CE ⇒ $R_e^x = (a/b)R_e^y = M/b$.
- Back-substitute $\Sigma F_x$ on BEG ⇒ $R_x^B = M/b$.
- Very systematic, no shortcuts, easy to adapt to any similar L-link problem.

Solution 2
- FBD of BEG ⇒ $R_e^y = -M/a$.
- Then identify CE as a two-force member → its force must lie along the line CE at angle $\theta = \tan^{-1}(b/a)$.
- Resolve that axial force into horizontal ⇒ $R_e^x = -M/b$.
- Back on BEG ⇒ $R_x^B = +M/b$.
- Exploits the two-force property elegantly but adds a trig step.

Solution 3, 5, 6, 7, 10
- Variations on the same theme as Solution 1, some choosing different moment centers or sign conventions, all arriving cleanly at M/b.
- They differ only in minor bookkeeping or in the order in which they write the equilibrium equations.

Why Solution 1 is my top pick

- Fully explicit FBDs of both members, writing out all $\Sigma F$ and $\Sigma M$ in a straightforward order.
- No hidden assumptions, no extra trig, and no reliance on remembering a special "two-force…" shortcut.
- Can be carried out in any coordinate orientation, with any kind of pinned joint, without losing clarity.
- Easy for a student or practicing engineer to check each step.

If you prefer to lean on the classical "two-force-member" shortcut, Solution 2 is almost as crisp, but it does introduce an angle-resolution step that in more complicated geometries could be error-prone.

Recommendation
Adopt the full two-body equilibrium approach in Solution 1 as your standard template.  It is the clearest, most systematic way to see exactly how the couple at G is transmitted through CE into a horizontal reaction at B, and it generalizes immediately to any members pinned together in L-shapes or other frames.



\end{appendices}

%
%
%
%
%
%
%


\end{document}